\documentclass[runningheads,a4paper]{llncs}

\usepackage{microtype}                 
\PassOptionsToPackage{warn}{textcomp}  
\usepackage{textcomp}                  
\usepackage{mathptmx}                  
\usepackage{times}                     
\usepackage{cite}                      
\usepackage{tabu}                      
\usepackage{booktabs}                  
\usepackage{amssymb}
\usepackage{bm}
\usepackage{amsmath}
\usepackage{amssymb}
\usepackage{xspace}
\usepackage{graphicx}
\usepackage{multirow}
\usepackage{color}
\usepackage{flushend}
\usepackage{nccmath}
\usepackage{makeidx}
\usepackage{hyperref}

\usepackage{textcomp}
\usepackage{fancyvrb}

\AtBeginEnvironment{subappendices}{%
\chapter*{Appendix}
\addcontentsline{toc}{chapter}{Appendices}
\counterwithin{figure}{section}
\counterwithin{table}{section}
}
\AtEndEnvironment{subappendices}{%
\counterwithout{figure}{section}
\counterwithin{figure}{chapter}
\counterwithout{table}{section}
\counterwithin{table}{chapter}
}

\usepackage{subfig}
\newcommand{\slfrac}[2]{\left.#1\middle/#2\right.}
\DeclareMathOperator*{\argmax}{arg\,max}
\DeclareMathOperator*{\argmin}{arg\,min}

\usepackage{url}
\urldef{\mailsa}\path|{alfred.hofmann, ursula.barth, ingrid.haas, frank.holzwarth,|
\urldef{\mailsb}\path|anna.kramer, leonie.kunz, christine.reiss, nicole.sator,|
\urldef{\mailsc}\path|erika.siebert-cole, peter.strasser, lncs}@springer.com|

\begin{document}
\begin{center}
This is the pre-submission version of the manuscript that was later edited and \\ published as a chapter in \textit{RGB-D Image Analysis and Processing}.

\vspace{2em}

DOI:

\url{http://doi.org/10.1007/978-3-030-28603-3_6}

\vspace{1em}

SpringerLink:

\url{http://link.springer.com/chapter/10.1007/978-3-030-28603-3_6}

\vspace{2em}

\textcopyright \ Springer Nature Switzerland AG 2019

\end{center}

\vspace{2em}

Cite as:
\begin{center}
    \fbox{\begin{minipage}{28em}
    Civera J., Lee S.H. (2019) RGB-D Odometry and SLAM. In: Rosin P., Lai YK., Shao L., Liu Y. (eds) RGB-D Image Analysis and Processing. Advances in Computer Vision and Pattern Recognition. Springer, Cham
\end{minipage}}
\end{center}

\vspace{2em}

BibTeX:
\begin{center}
\begin{minipage}{28.5em}
\begin{Verbatim}[frame=single]
@Inbook{Civera2019,
author="Civera, Javier
and Lee, Seong Hun",
editor="Rosin, Paul L.
and Lai, Yu-Kun
and Shao, Ling
and Liu, Yonghuai",
title="RGB-D Odometry and SLAM",
bookTitle="RGB-D Image Analysis and Processing",
year="2019",
publisher="Springer International Publishing",
address="Cham",
pages="117--144",
isbn="978-3-030-28603-3",
doi="10.1007/978-3-030-28603-3_6",
url="https://doi.org/10.1007/978-3-030-28603-3_6"}
\end{Verbatim}
\end{minipage}
\end{center}

\pagenumbering{gobble}
\newpage
\pagenumbering{arabic}

\title{RGB-D Odometry and SLAM}

\titlerunning{Chapter 6 of RGB-D Image Analysis and Processing}

%
%
\author{Javier Civera%
\and Seong Hun Lee\thanks{This work was partially supported by the Spanish government (project PGC2018-096367-B-I00) and the Aragón regional government (Grupo DGA-T45 17R/FSE).  }}
\authorrunning{Chapter 6 of RGB-D Image Analysis and Processing}

\institute{I3A, Universidad de Zaragoza, Spain,\\ \email{\{jcivera, seonghunlee\}@unizar.es}}

%
%

\toctitle{Lecture Notes in Computer Science}
\tocauthor{Authors' Instructions}
\maketitle

\begin{abstract}

The emergence of modern RGB-D sensors had a significant impact in many application fields, including robotics, augmented reality (AR) and 3D scanning. 
They are low-cost, low-power and low-size alternatives to traditional range sensors such as LiDAR.
Moreover, unlike RGB cameras, RGB-D sensors provide the additional depth information that removes the need of frame-by-frame triangulation for 3D scene reconstruction. 
These merits have made them very popular in mobile robotics and AR, 
where it is of great interest to estimate ego-motion and 3D scene structure.
Such spatial understanding can enable robots to navigate autonomously without collisions and allow users to insert virtual entities consistent with the image stream. 
In this chapter, we review common formulations of odometry and Simultaneous Localization and Mapping (known by its acronym SLAM) using RGB-D stream input. 
The two topics are closely related, as the former aims to track the incremental camera motion with respect to a local map of the scene, and the latter to jointly estimate the camera trajectory and the global map with consistency. 
In both cases, the standard approaches minimize a cost function using nonlinear optimization techniques. 
This chapter consists of three main parts:
In the first part, we introduce the basic concept of odometry and SLAM and motivate the use of RGB-D sensors.  
We also give mathematical preliminaries relevant to most odometry and SLAM algorithms.
In the second part, we detail the three main components of SLAM systems: camera pose tracking, scene mapping and loop closing.
For each component, we describe different approaches proposed in the literature.
In the final part, we provide a brief discussion on advanced research topics with the references to the state-of-the-art.
\end{abstract}

\section{Introduction: SLAM\index{SLAM}, Visual SLAM\index{visual SLAM} and RGB-D Sensors}

Visual Odometry\index{visual odometry} and Visual Simultaneous Localization and Mapping -- from here on referred to as their respective acronyms VO and VSLAM -- are two tightly related topics that aim to extract 3D information from streams of visual data in real-time. Specifically, the goal of VO is to estimate the incremental motion (\emph{i.e.}, translation and rotation) of the camera as it moves. The goal of Visual SLAM is more ambitious: To estimate a globally consistent map of the scene and the camera trajectory with respect to it. 


In the robotics research community, SLAM is considered as a fundamental capability for autonomous robots. See \cite{durrant2006simultaneous,bailey2006simultaneous} for an illustrative tutorial covering the earliest approaches, and \cite{cadena2016past} for a recent survey outlining the state-of-the-art and the most relevant future directions. While the early pioneering works on SLAM mainly used laser scanners (\emph{e.g.,} \cite{castellanos1999spmap}), the field rapidly pivoted to cameras for several reasons. Among them were the progress of computer vision algorithms and improved processors, as well as the camera's low cost, size and power consumption.


\begin{figure}[ht!]
\centering
\subfloat[RGBDSLAM \cite{endres20143}]{
\centering
\includegraphics[width=0.3\textwidth]{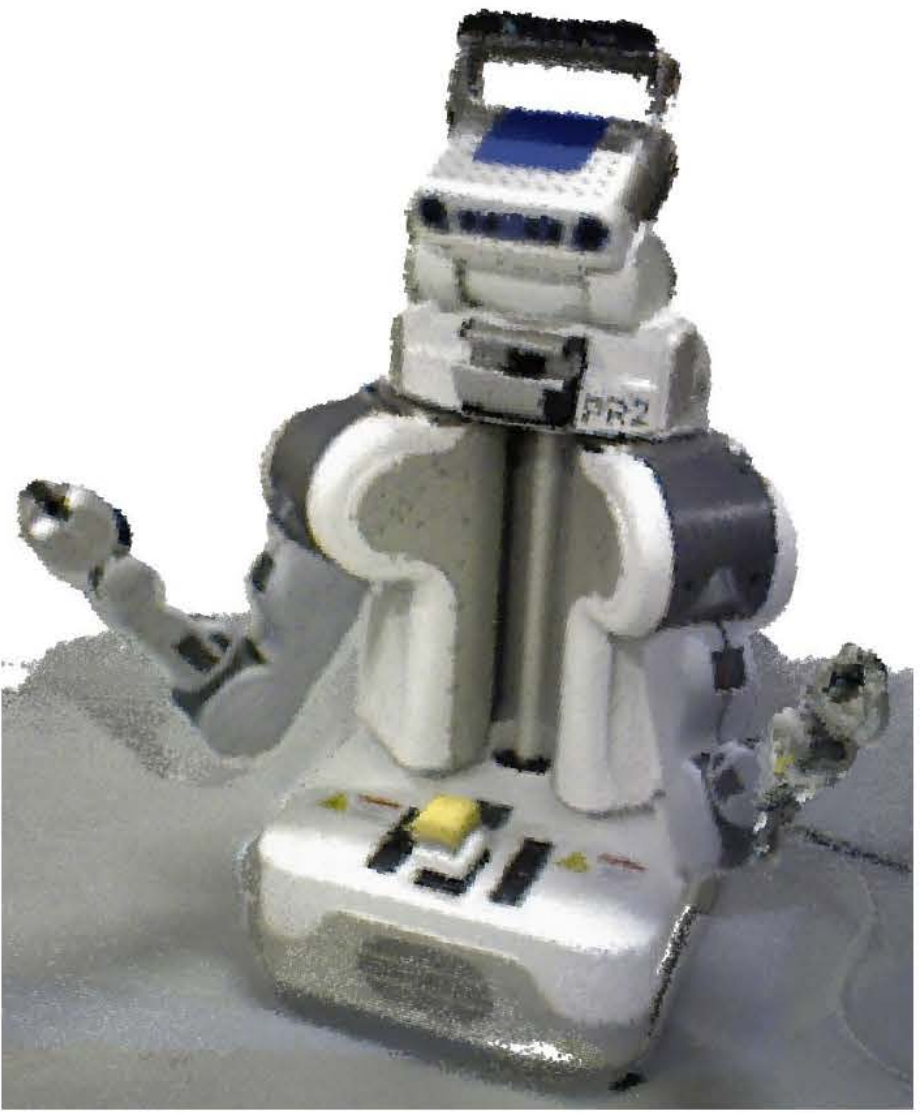}
\label{fig:rgbdslam}
}
\subfloat[ORB-SLAM2 \cite{mur2017orb}]{
\centering
\includegraphics[width=0.43\textwidth]{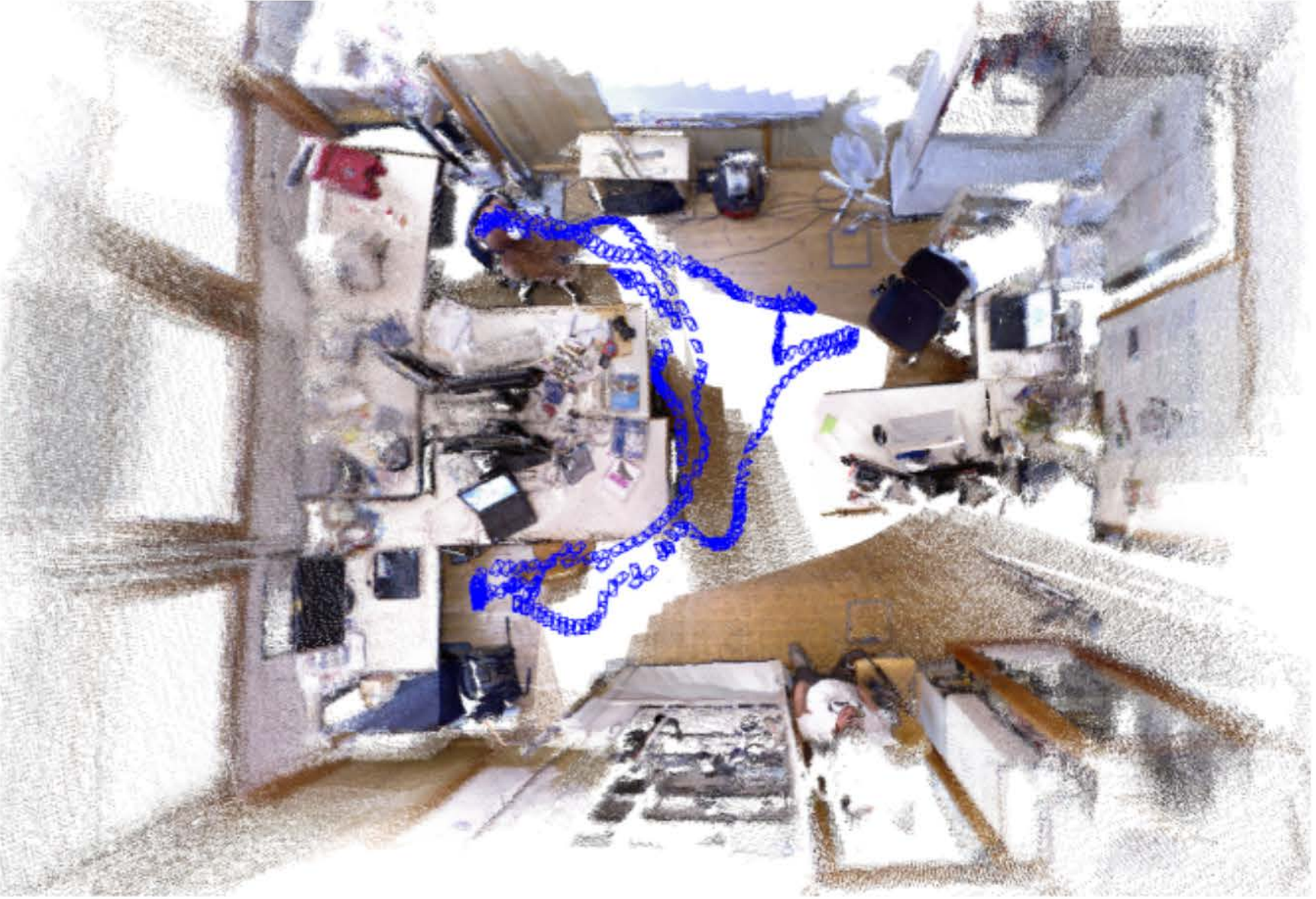}}
\label{fig:orbslam2}
~\\
\subfloat[ElasticFusion \cite{whelan2016elasticfusion}]{
\centering
\includegraphics[width=0.48\textwidth]{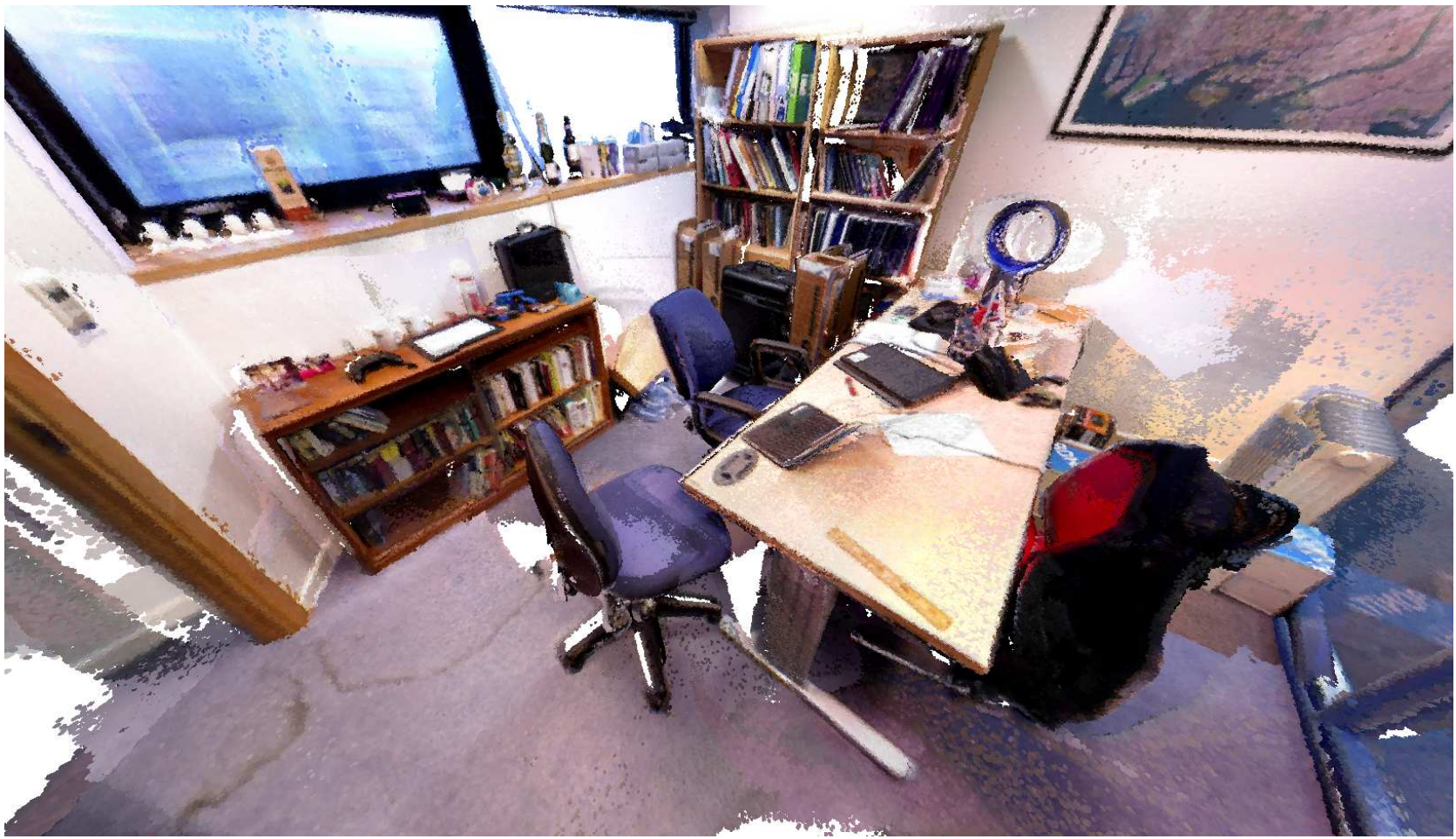}
\label{fig:elasticfusion}
}
\subfloat[RGBDTAM \cite{concha2017rgbdtam}]{
\centering
\includegraphics[width=0.43\textwidth]{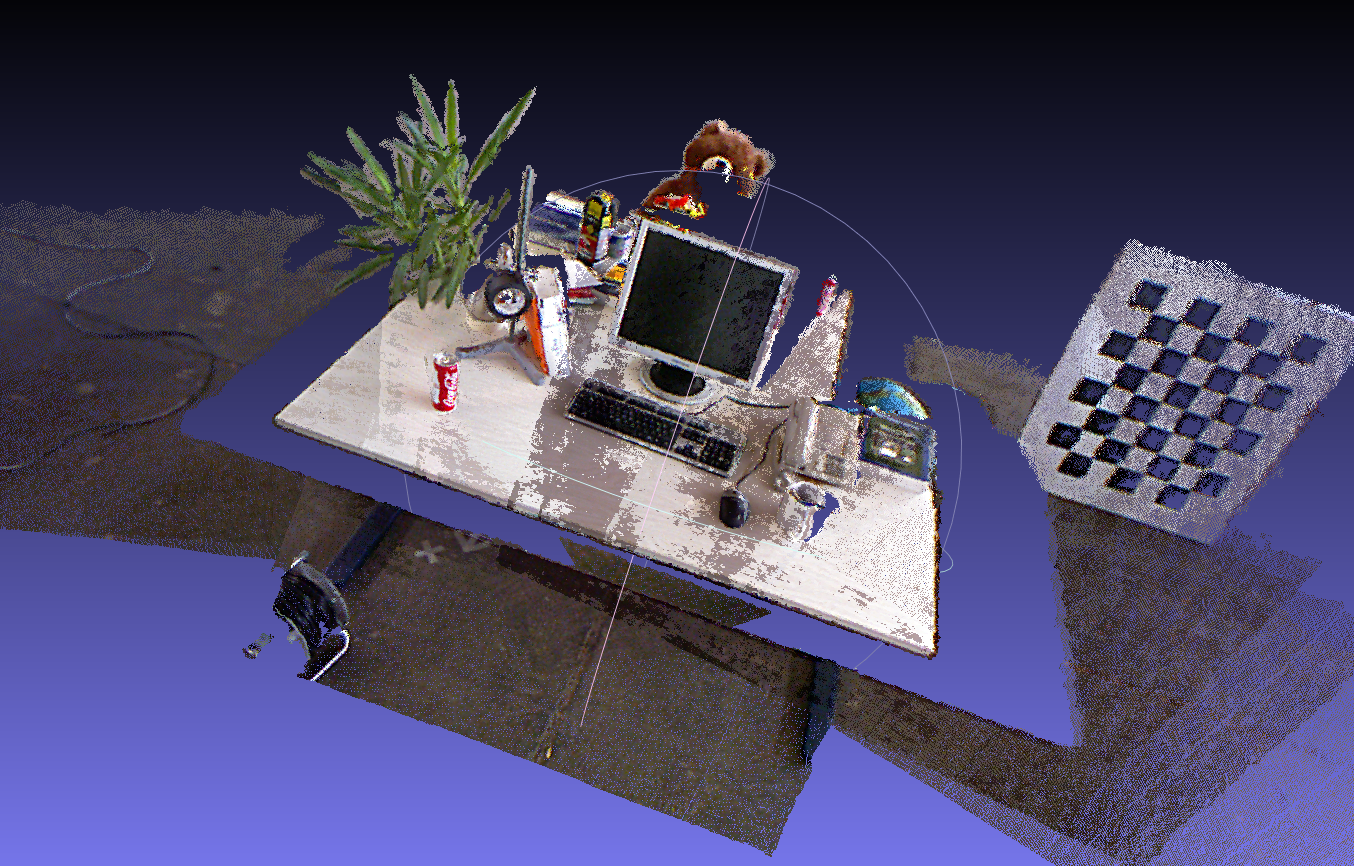}
\label{fig:rgbdtam}
}
\caption{State-of-the-art RGB-D SLAM systems.}
\label{fig:rgbd-slam}
\end{figure}

Most visual SLAM methods have been traditionally based on low-level feature matching and multiple view geometry. 
This introduces several limitations to monocular SLAM. 
For example, a large-baseline motion is needed to generate sufficient parallax for reliable depth estimation; and the scale is unobservable. 
This can be partially alleviated by including additional sensors (\emph{e.g.}, stereo cameras \cite{pire2017s}, inertial measurement units (IMUs) \cite{concha2016visual}, sonar \cite{engel2014scale}) or the prior knowledge of the system \cite{lee2018stability} or the scene \cite{tateno2017cnn}.
Another challenge is the dense reconstruction of low texture areas \cite{concha2014manhattan}.
Although recent approaches using deep learning (\emph{e.g.}, \cite{bloesch2018codeslam,zhou2018deeptam}) have shown impressive results in this direction, more research is needed regarding their cost and dependence on the training data \cite{facil2019camconvs}.

RGB-D sensors provide a practical hardware-based alternative to the challenges and limitations mentioned above. Their availability at low cost has facilitated many robotics and AR applications in the last decade. Intense research endeavors have produced numerous robust algorithms and real-time systems. Figure \ref{fig:rgbd-slam} shows several reconstruction examples from the state-of-the-art systems. Today, RGB-D cameras stand out as one of the preferred sensors for indoor applications in robotics and AR; and their future looks promising either on their own or in combination with additional sensors.

In this chapter, we will cover several state-of-the-art RGB-D odometry and SLAM algorithms. Our goal is to focus on the basic aspects of geometry and optimization, highlighting relevant aspects of the most used formulations and pointing to the most promising research directions. The reader should be aware that, as a consequence of condensing a vast array of works and presenting the basics in a homogeneous and easy-to-follow manner, some individual works might present slight variations from the formulation presented here. In general, we sacrificed extending ourselves over particular details in favour of a clearer overview of the field. 

The rest of the chapter is organized as follows. Section \ref{sec:rgbdoslampipelines} will give an overview on the most usual VO and VSLAM pipeline.
Section \ref{sec:notation} will introduce the notation used throughout the rest of the chapter. Section \ref{sec:motion} will cover the algorithms for tracking the camera pose, Section \ref{sec:mapping} the algorithms for the estimation of the scene structure, and Section \ref{sec:loopclosure} the loop closure algorithms. Section \ref{sec:references} will refer to relevant scientific works and research lines that were not covered in the previous sections. Finally, Section \ref{sec:conclusions} contains the conclusions and Section \ref{sec:resources} provides links to some of the most relevant online resources, mainly the state-of-the-art open-source software and public benchmark datasets. 


\section{The Visual Odometry and SLAM pipelines}
\label{sec:rgbdoslampipelines}

The pipelines of RGB-D Odometry and SLAM have many components in common. Here, we will give a holistic view of the building blocks of standard implementations, highlighting their connections and introducing the terminology. 

The seminal work of Klein and Murray \cite{klein2007parallel} proposed the architecture that is used in most visual odometry and SLAM systems nowadays. Such architecture was later refined in papers like \cite{strasdat2010scale,mur2015orb,engel2018direct} among others. Basically, the idea is to partition the processing into two (or more) parallel threads: one thread tracks the camera pose in real time at video rate, and the rest update several levels of scene representations at lower frequencies (in general, the larger and/or more complex the map, the lower the frequency of update). 

\begin{figure}[ht!]
\centering
\includegraphics[width=0.9\textwidth]{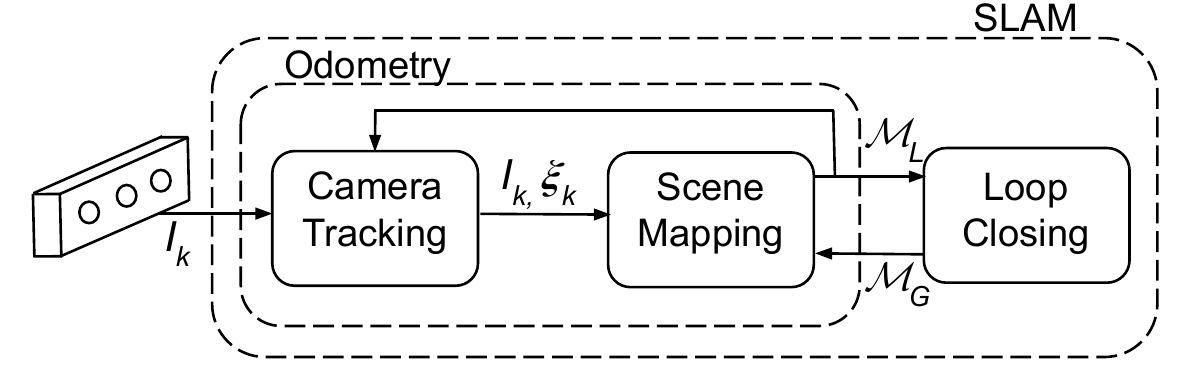}
\caption{High-level overview of VO and VSLAM systems. $I_k$: $k^{th}$ RGB-D image, $\xi_k$: $k^{th}$ camera pose, $\mathcal{M}_L$ and $\mathcal{M}_G$: the local and the global map.}
\label{fig:pipeline}
\end{figure}

Fig. \ref{fig:pipeline} illustrates a simple Tracking and Mapping architecture for RGB-D Odometry and SLAM that we will use in this chapter. The camera tracking thread estimates the camera motion $\mathbf{\xi}_k$ at time $k$ given the current frame $I_k$ and a local map $\mathcal{M}_L$. $\mathcal{M}_L$ is estimated from a set of keyframes summarizing the sequence. If SLAM is the aim, a globally consistent map $\mathcal{M}_G$ is estimated by means of loop closure and global optimization. In more detail:

\begin{itemize}

    \item {\bf Camera tracking: }The camera tracking thread estimates the incremental camera motion. The most simple approach is to use the frame-to-frame constraints (\emph{e.g.}, \cite{kerl2013robust,gutierrez2016dense}). 
    This is in fact inevitable when bootstrapping the system from the first two views. 
    However, after initialization, it is quite usual to use more than two views in order to achieve higher accuracy. 
    In this case, the standard approach is to estimate the camera motion using map-to-frame constraints with respect to a local map built from the past keyframes (see the next paragraph). 
    
    \item {\bf Scene Mapping: } Mapping approaches vary, depending on the application. Volumetric mapping discretizes the scene volume into voxels and integrates the information from the RGB-D views (\emph{e.g.}, \cite{newcombe2011kinectfusion,whelan2015real}). Point-based mapping performs a nonlinear optimization of camera poses and points (\emph{e.g.}, \cite{mur2017orb}).
    In the case of VO, the map is local and is estimated from a sliding window containing a selection of the last frames (\emph{e.g.}\cite{wang2014rgbd,engel2018direct}). 
    In the case of VSLAM, the map is estimated from a set of keyframes representative of the visited places.
    
    \item {\bf Loop Closing: } In both odometry and SLAM drift is accumulated in purely exploratory trajectories. Such drift can be corrected if a place is revisited, using approaches denoted as loop closure. First, the place is recognized by its visual appearance (loop detection), and then the error of the global map is corrected (loop correction) \cite{mur2017orb,gutierrez2018rgbid,whelan2015real}).
    
\end{itemize}

\section{Notation and Preliminaries}
\label{sec:notation}

\subsection{Geometry and Sensor Model\index{camera model}}
\label{sec:geometry}

We denote an RGB-D input as $I:\Omega \mapsto \mathbb{R}^4$, where $\Omega \subset \mathbb{R}^2$ is the image plane of width $w$ and height $h$. 
We represent the pixel coordinates as a 2D vector $\mathbf{p} = \left( u, v \right)^\top$ and the corresponding homogeneous coordinates as  $\tilde{\mathbf{p}} = (\tilde{u}, \tilde{v}, \lambda)^\top$.
Each pixel has RGB color and depth value, \emph{i.e.}, $I\left( u, v \right) = \left( r, g, b, d \right)^\top$. 
The depth channel is denoted as $D:\Omega \mapsto \mathbb{R}$, and the access to it as $D\left( u, v \right) = d$.
The Euclidean coordinates of a 3D point $k$ in some reference frame $i$ (be it a camera or the world reference) are denoted by $\mathbf{P}^i_k = \left( X^i_k, Y^i_k, Z^i_k \right)^\top$ or $\mathbf{\tilde P}^i_k = \left( \lambda X^i_k, \lambda Y^i_k, \lambda Z^i_k, \lambda \right)^\top$ in homogeneous coordinates. 
These two coordinates are related by the dehomogenization operation:
$\mathbf{P}^i_k=\pi_\text{3D}(\mathbf{\tilde{P}}_k^i)$.
Inversely, the homogenization is denoted by $\pi^{-1}_\text{3D}(\mathbf{P}^i_k) := \left( X^i_k, Y^i_k, Z^i_k, 1 \right)^\top$.


\begin{figure}[t]
\centering
\includegraphics[width=0.6\textwidth]{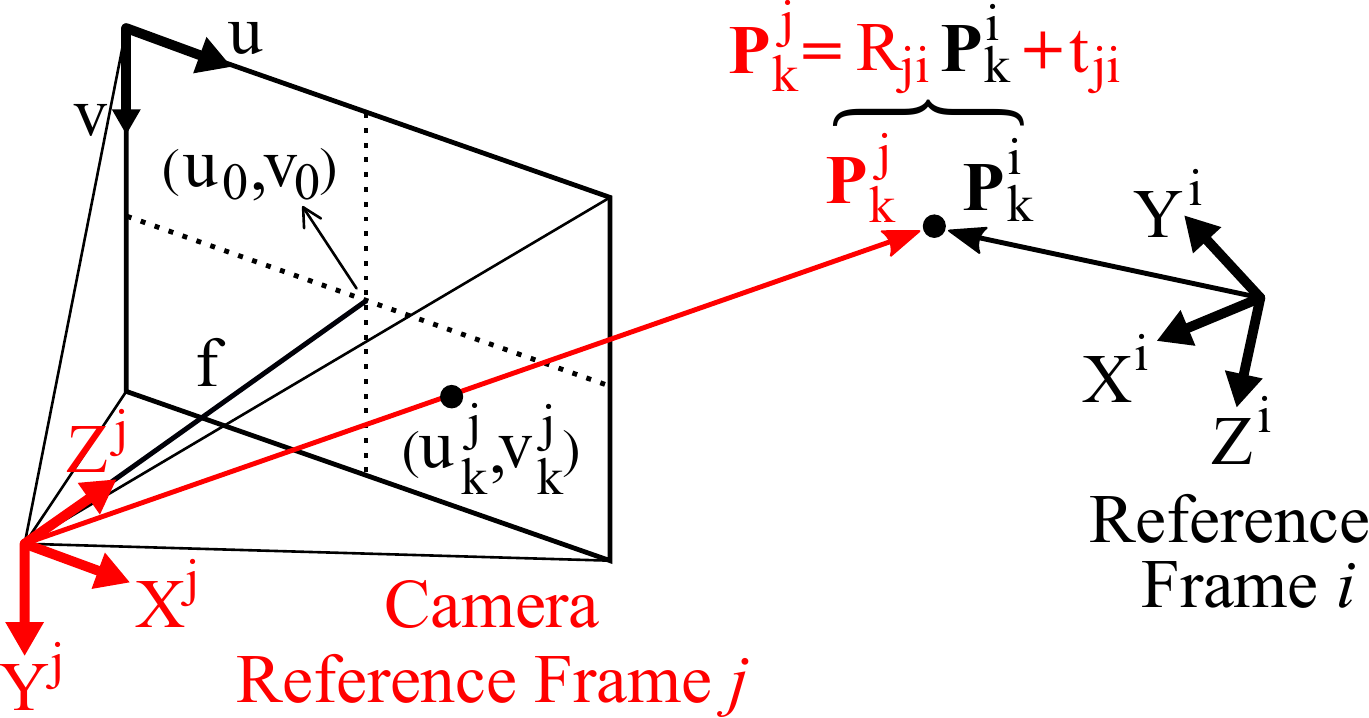}
\caption{The transformation of point $k$ from reference frame $i$ to camera reference frame $j$, and its projection onto the image plane using the pinhole camera model. }
\label{fig:geometry_model}
\end{figure}
The pose of camera $j$ with respect to reference frame $i$ is defined by the transformation $\mathtt{T}_{ji} = \begin{bmatrix} \mathtt{R}_{ji} & \mathbf{t}_{ji} \\ \mathtt{0} & 1 \end{bmatrix} \in \ensuremath{\mathrm{SE}(3)}\xspace$, $\mathtt{R}_{ji}\in \ensuremath{\mathrm{SO}(3)}\xspace$, $\mathbf{t}_{ji} \in \mathbb{R}^3$.
The rotation matrix $\mathtt{R}_{ji}$ and translation vector $\mathbf{t}_{ji}$ are defined such that the transformation of point $\mathbf{P}^i_k$ in reference frame $i$ to the $j^{th}$ camera reference frame is
\begin{equation}
\mathbf{\tilde P}^j_k = \mathtt{T}_{ji} \mathbf{\tilde P}_k^i; \ \ \mathbf{P}^j_k = \mathtt{R}_{ji} \mathbf{P}^i_k + \mathbf{t}_{ji}.
\label{eq:transfwtoj}
\end{equation}
Likewise, $\tilde{\mathbf{P}}^i_k$ can be obtained from $\tilde{\mathbf{P}}^i_k$ and $\mathtt{T}_{ji}$ with the inverse operation:
\begin{equation}
    \mathbf{\tilde{P}}^{i}_k=\mathtt{T}_{ji}^{-1}\mathbf{\tilde{P}}^j_k; \ \ \mathbf{P}^{i}_k = \mathtt{R}_{ji}^\top\left(\mathbf{P}^j_k-\mathbf{t}_{ji}\right).
\end{equation}
As illustrated in Fig. \ref{fig:geometry_model}, we adopt the standard pinhole model for the projection onto the image plane.
First, the 3D point $\mathbf{P}^i_k$ is transformed to the camera frame $j$ using equation \ref{eq:transfwtoj}. 
The homogeneous coordinates of the projection in the image space are given by
\begin{equation}
\label{eq:projection}
    \tilde{\mathbf{p}}^j_k = (\tilde{u}^j_k, \tilde{v}^j_k, \lambda)^\top=\mathtt{K}\mathbf{P}^j_k=\mathtt{K}(\mathtt{R}_{ji} \mathbf{P}^i_k + \mathbf{t}_{ji}) \quad \text{with} \quad \mathtt{K}=\begin{bmatrix}
f_x & 0 & u_0 \\
0 & f_y & v_0 \\
0 & 0 & 1 
\end{bmatrix},
\end{equation}
where $\mathtt{K}$ is the calibration matrix containing the coordinates of the principal point $(u_0, v_0)^\top$ and the focal lengths $(f_x, f_y)=(fm_x, fm_y)$.
Here, $(m_x, m_y)$ denotes the number of pixels per unit distance in image coordinates in the $x$ and $y$ directions.
The pixel coordinates are finally obtained by dehomogenization: $\mathbf{p}^j_k=(u^j_k, v^j_k)^\top=\pi_\text{2D}(\tilde{\mathbf{p}}^j_k)=(\slfrac{\tilde{u}^j_k}{\lambda}, \slfrac{\tilde{v}^j_k}{\lambda})^\top$. 
The inverse operation (\emph{i.e.}, homogenization) is denoted by $\pi_\text{2D}^{-1}(\mathbf{p}^j_k):=(u^j_k, v^j_k, 1)^\top$.

Now, let reference frame $i$ be another camera reference frame.
Then, the \textit{reprojection} of 2D point $\mathbf{p}^i_k$ in frame $i$ to frame $j$ is defined as the following three-step operation:
\begin{enumerate} \itemsep0em
    \item Backproject $\mathbf{p}^i_k$ with the measured depth $d^i_k$ to estimate the 3D point $\mathbf{P}^i_k$ in frame~$i$:
    \begin{equation}
    \label{eq:backprojection}
        {\mathbf{P}^{i}_k}'=d^i_k\frac{\mathtt{K}^{-1}\pi^{-1}_\text{2D}
        \left(\mathbf{p}^i_k\right)}{\big\lVert\mathtt{K}^{-1}\pi^{-1}_\text{2D}\left(\mathbf{p}^i_k\right)\big\rVert}.
    \end{equation}
    \item Transform this estimate from frame $i$ to frame $j$:
    \begin{equation}
        \mathbf{P}^{ji}_k = \pi_\text{3D}\left(\mathtt{T}_{ji}\pi^{-1}_\text{3D}
        \left({\mathbf{P}^{i}_k}'\right)\right); \ 
        \mathbf{P}^{ji}_k = \mathtt{R}_{ji}{\mathbf{P}^{i}_k}'+\mathbf{t}_{ji}
    \end{equation}
    (Notice that we use the superscript ${ji}$ instead of $j$ to distinguish the ground-truth in frame $j$.) 
    \item Project the resulting 3D point to obtain its pixel coordinates in frame $j$.
    \begin{equation}
        \mathbf{p}^{ji}_k = \pi_\text{2D}\left(\mathtt{K}\mathbf{P}^{ji}_k\right).
    \end{equation}
\end{enumerate}
Altogether, the reprojection of point $\mathbf{p}^i_k$ to frame $j$ is defined as follows:
\begin{equation}
\label{eq:reprojection}
    \mathbf{p}^{ji}_{k}\left(\mathbf{p}^i_k, \ d^i_k,  \mathtt{T}_{ji}\right)
    =
    \pi_\text{2D}\left(\mathtt{K}\left(\frac{d^i_k\mathtt{R}_{ji}\mathtt{K}^{-1}\pi^{-1}_\text{2D}\left(\mathbf{p}^i_k\right)}{\big\lVert\mathtt{K}^{-1}\pi^{-1}_\text{2D}\left(\mathbf{p}^i_k\right)\big\rVert}+\mathbf{t}_{ji}\right)\right).
\end{equation}


\subsection{Nonlinear Optimization\index{nonlinear optimization}}
\label{sec:optimization}

Most state-of-the-art VO and VSLAM methods rely heavily on nonlinear optimization in order to estimate the state vector $\mathbf{x}$ (\emph{e.g.}, containing the camera poses and 3D map points) from a set of noisy measurements $\mathbf{z} = \{\mathbf{z}_1, \mathbf{z}_2, \hdots \}$ (\emph{e.g.}, image correspondences or pixel intensities).


According to Bayes' theorem, the following equation describes the conditional probability of the state $p(\mathbf{x}|\mathbf{z})$ given the measurement model $p(\mathbf{z}|\mathbf{x})$ and the prior over the state $p(\mathbf{x})$:
\begin{equation}
\label{eq:bayes}
    p(\mathbf{x}|\mathbf{z}) = \frac{p(\mathbf{z}|\mathbf{x})p(\mathbf{x})}{p(\mathbf{z})}
\end{equation}
Our aim is then to find the state $\mathbf{x}$ that maximizes this probability.
This is called the Maximum a Posteriori (MAP) problem, and the solution corresponds to the mode of the posterior distribution:
\begin{equation}
\label{eq:MAP}
    \mathbf{x}_{MAP} = \argmax_{\mathbf{x}} p(\mathbf{x}|\mathbf{z}) = \argmax_{\mathbf{x}} \frac{p(\mathbf{z}|\mathbf{x})p(\mathbf{x})}{p(\mathbf{z})}
\end{equation}

Modern VSLAM and VO methods are based on smoothing and often assume a uniform prior $p(\mathbf{x})$.
The normalization constant $p(\mathbf{z})$ does not depend on the state either.
Therefore, we can drop $p(\mathbf{x})$ and $p(\mathbf{z})$ from \eqref{eq:MAP}, turning the problem into the Maximum Likelihood Estimation (MLE).
Assuming the independence between the measurements, this means that \eqref{eq:MAP} becomes
\begin{equation}
    \mathbf{x}_{MAP}= \mathbf{x}_{MLE}=\argmax_{\mathbf{x}} p(\mathbf{z}|\mathbf{x}) = \argmax_{\mathbf{x}} \displaystyle \prod_{k} p(\mathbf{z}_k|\mathbf{x}).
\end{equation}
Suppose that the measurement model is given by $\mathbf{z}_k = \mathbf{h}_k(\mathbf{x}) + \mathbf{\delta}_k$, where $\mathbf{\delta}_k \sim \mathcal{N}(\mathbf{0},\mathtt{\Omega}_k)$. The conditional distribution of the individual measurements is then $p(\mathbf{z}_k|\mathbf{x}) \sim \mathcal{N}(\mathbf{h}_k(\mathbf{x}),\mathtt{\Omega}_k)$. Maximizing, for convenience, the $\log$ of the conditionals leads to

\begin{align}
    &\mathbf{x}_{MAP} = \argmax_{\mathbf{x}} \log(\displaystyle \prod_{k} p(\mathbf{z}_k|\mathbf{x})) =  \argmax_{\mathbf{x}} \sum_{k} \log( p(\mathbf{z}_k|\mathbf{x})) \nonumber\\
    &= \argmax_{\mathbf{x}} \sum_{k} \log( \exp(-\frac{1}{2}(\mathbf{z}_k-\mathbf{h}_k(\mathbf{x}))^\top \mathtt{\Omega}_k^{-1} (\mathbf{z}_k-\mathbf{h}_k(\mathbf{x})))) = \argmin_{\mathbf{x}} \sum_{k} ||\mathbf{r}_k(\mathbf{x})||_{\mathtt{\Omega}_k}^2, \label{eq:wls}
\end{align}
\noindent where $||\mathbf{r}_k(\mathbf{x})||_{\mathtt{\Omega}_k}=\sqrt{(\mathbf{z}_k-\mathbf{h}_k(\mathbf{x}))^\top \mathtt{\Omega}_k^{-1} (\mathbf{z}_k-\mathbf{h}_k(\mathbf{x}))}$ is called the \textit{Mahalanobis} distance.
As $\mathbf{h}_k(\mathbf{x})$ is typically nonlinear, we solve \eqref{eq:wls} using an iterative method.
A standard approach is to use the Gauss-Newton\index{Gauss-Newton} algorithm described as follows:
\begin{enumerate}\itemsep0em
    \item Make an initial guess $\breve{\mathbf{x}}$.
    \item Linearize \eqref{eq:wls} using the Taylor approximation at $\breve{\mathbf{x}}$.
    \item Compute the optimal increment $\Delta\mathbf{x}^*$ that minimizes the linearized cost function.
    \item Update the state: $\breve{\mathbf{x}}\leftarrow\breve{\mathbf{x}}+\Delta\mathbf{x}^*$.
    \item Iterate the Step 2 -- 4 until convergence.
\end{enumerate}
The Taylor approximation in Step 2 gives
\begin{equation}
    \mathbf{h}_k(\breve{\mathbf{x}}+\Delta\mathbf{x})\approx \mathbf{h}_k(\breve{\mathbf{x}})+\mathbf{J}_k\Delta\mathbf{x} \quad \text{with} \quad \mathbf{J}_k=\frac{\partial \mathbf{h}_k(\mathbf{x})}{\partial \mathbf{x}}\biggr\rvert_{\breve{\mathbf{x}}}.
\end{equation}
This allows us to approximate $||\mathbf{r}_k(\breve{\mathbf{x}}+\Delta\mathbf{x})||_{\mathtt{\Omega}_k}^2$ as
\begin{align}
    ||\mathbf{r}_k(\breve{\mathbf{x}}+\Delta\mathbf{x})||_{\mathtt{\Omega}_k}^2
    &=
    (\mathbf{z}_k-\mathbf{h}_k(\breve{\mathbf{x}}+\Delta\mathbf{x}))^\top \mathtt{\Omega}_k^{-1} (\mathbf{z}_k-\mathbf{h}_k(\breve{\mathbf{x}}+\Delta\mathbf{x}))\\
    &\approx
    (\mathbf{z}_k-\mathbf{h}_k(\breve{\mathbf{x}})-\mathbf{J}_k\Delta\mathbf{x})^\top \mathtt{\Omega}_k^{-1} (\mathbf{z}_k-\mathbf{h}_k(\breve{\mathbf{x}})-\mathbf{J}_k\Delta\mathbf{x})\\
    &=
    \Delta\mathbf{x}^\top\mathbf{J}_k^\top\mathtt{\Omega}_k^{-1}\mathbf{J}_k\Delta\mathbf{x}
    +
    (\mathbf{z}_k-\mathbf{h}_k(\breve{\mathbf{x}}))^\top\mathtt{\Omega}_k^{-1}(\mathbf{z}_k-\mathbf{h}_k(\breve{\mathbf{x}})) \nonumber\\
    & \quad -2(\mathbf{z}_k-\mathbf{h}_k(\breve{\mathbf{x}}))^\top\mathtt{\Omega}_k^{-1}\mathbf{J}_k\Delta\mathbf{x}.
\end{align}
Now, taking the derivative of $\sum_k ||\mathbf{r}_k(\breve{\mathbf{x}}+\Delta\mathbf{x})||_{\mathtt{\Omega}_k}^2$ with respect to $\Delta\mathbf{x}$ and setting it to zero, we obtain the optimal increment in the following form:
\begin{equation}
    \Delta\mathbf{x}^*=-\underbrace{\left[\sum_k \mathbf{J}_k^\top\mathtt{\Omega}_k^{-1}\mathbf{J}_k\right]^{-1}}_{\displaystyle\mathbf{H}^{-1}}\underbrace{\sum_k \mathbf{J}_k^\top\mathtt{\Omega}_k^{-1}(\mathbf{h}_k(\breve{\mathbf{x}})-\mathbf{z}_k)}_{\displaystyle\mathbf{b}}.
\end{equation}
The Levenberg-Marquardt algorithm, a variant of the Gauss-Newton method, includes a non-negative damping factor $\lambda$ in the update step:
\begin{equation}
    \Delta\mathbf{x}^*=-\left(\mathbf{H}+\lambda \ \text{diag}(\mathbf{H})\right)^{-1}\mathbf{b},
\end{equation}
where $\lambda$ is increased when the cost function reduces too slowly, and vice versa.
For more details on the adjustment rule, see \cite{madsen2004methods}.

Since the least squares problems are very sensitive to outliers, a common practice is to adopt a robust weight function that downweights large errors: 
\begin{equation}
    \mathbf{x}_\text{robust} = \argmin_{\mathbf{x}} \sum_{k} \omega\left(||\mathbf{r}_k(\mathbf{x})||_{\mathtt{\Omega}_k}\right) ||\mathbf{r}_k(\mathbf{x})||_{\mathtt{\Omega}_k}^2.
\end{equation}

To solve this problem iteratively, it is usually assumed that the weights are dependent on the residual at the previous iteration, which turns the problem into a standard weighted least squares at each iteration.
This technique is called the iteratively reweighted least squares (IRLS).
The readers are referred to \cite{huber2011robust,zhang1997parameter} for more details on the robust cost functions and \cite{barfoot2017state} for in-depth study of state estimation for robotics.

\subsection{Lie Algebras\index{Lie Algebras}}

Standard optimization techniques assume that the state belongs to a Euclidean vector space. This does not hold for 3D rotation matrices $\mathtt{R}$, belonging to the special orthogonal group $\ensuremath{\mathrm{SO}(3)}\xspace$, or for 6 degrees-of-freedom (DoF) rigid body motions $\mathtt{T}$, belonging to the special Euclidean group $\ensuremath{\mathrm{SE}(3)}\xspace$. 
In both cases, state updates have to be done in the tangent space of $\ensuremath{\mathrm{SO}(3)}\xspace$ and $\ensuremath{\mathrm{SE}(3)}\xspace$ at the identity, which are denoted as $\ensuremath{\mathfrak{so}(3)}\xspace$ and $\ensuremath{\mathfrak{se}(3)}\xspace$. 
Elements of the tangent space $\ensuremath{\mathfrak{so}(3)}\xspace$ and $\ensuremath{\mathfrak{se}(3)}\xspace$ can be represented as vector $\bm{\omega} \in \mathbb{R}^3$ and $\bm{\xi}=[\bm{\omega}, \bm{\nu}]^\top \in \mathbb{R}^6$, respectively.

The \emph{hat} operator $(\cdot)^\wedge$ converts $\bm{\omega} \in \mathbb{R}^3$ to the space of skew symmetric matrices of the Lie algebra and its inverse is denoted by the \emph{vee} operator $(\cdot)^\vee$:
\begin{equation}
    \bm{\omega}^\wedge
    = 
    \begin{bmatrix}
    \omega_x\\ \omega_y\\ \omega_z
    \end{bmatrix}^\wedge
    =
    \begin{bmatrix}
    0 & -\omega_z & \omega_y\\
    \omega_z & 0 & -\omega_x\\ 
    -\omega_y & \omega_x & 0
    \end{bmatrix}\in\ensuremath{\mathfrak{so}(3)} \quad \text{and} \quad \left(\bm{\omega}^\wedge\right)^\vee=\bm{\omega}\in\mathbb{R}^3.
\end{equation}
We denote the exponential and logarithmic mapping between $\ensuremath{\mathfrak{se}(3)}\xspace$ and $\ensuremath{\mathrm{SE}(3)}\xspace$ by $\mathrm{exp}_\text{SE(3)}(\bm{\xi})$ and $\mathrm{log_\text{SE3}(\mathtt{T})}$, respectively:
\begin{equation}
\label{eq:exp_se3}
    \mathrm{exp}_\mathrm{SE(3)}(\bm{\xi}):=
    \begin{bmatrix}
    \mathrm{exp}(\bm{\omega}^\wedge) & \mathtt{V}\bm{\nu}\\
    0 & 1
    \end{bmatrix}
    =\begin{bmatrix}
        \mathtt{R} & \mathbf{t}\\
        0 & 1
    \end{bmatrix}
    =\mathtt{T}
    \in \mathrm{SE}(3),
\end{equation}
where
\begin{equation}
    \mathrm{exp}(\bm{\omega}^\wedge)=
    \mathtt{I}_{3\times3}+\frac{\sin{\left(\lVert\bm{\omega}\rVert\right)}}{\lVert\bm{\omega}\rVert}\bm{\omega}^\wedge+\frac{1-\cos{\left(\lVert\bm{\omega}\rVert\right)}}{\lVert\bm{\omega}\rVert^2}\left(\bm{\omega}^\wedge\right)^2
\end{equation}
and
\begin{equation}
    \mathtt{V}
    =
    \mathtt{I}_{3\times3}
    +\frac{1-\cos{\lVert\bm{\omega}\rVert}}{\lVert\bm{\omega}\rVert^2}\bm{\omega}^\wedge
    +\frac{\lVert\bm{\omega}\rVert-\sin{\left(\lVert\bm{\omega}\rVert\right)}}{\lVert\bm{\omega}\rVert^3}\left(\bm{\omega}^\wedge\right)^2.
\end{equation}
From \eqref{eq:exp_se3}, the logarithm map can be obtained:
\begin{equation}
    \mathrm{log}_\text{SE(3)}(\mathtt{T}):=
    \begin{bmatrix}
    \left(\log\mathtt{R}\right)^\vee \\
    \mathtt{V}^{-1}\mathbf{t}
    \end{bmatrix},
\end{equation}
where 
\begin{equation}
    \log\mathtt{R} =\frac{\theta}{2\sin{\theta}}\left(\mathtt{R}-\mathtt{R}^\top\right) \quad \text{with} \quad \theta=\cos^{-1}\left(\frac{\mathrm{trace}(\mathtt{R})-1}{2}\right).
\end{equation}

For optimization purposes, rigid body transformations can be conveniently represented as $\mathrm{exp}_\mathrm{SE(3)}(\Delta\bm{\xi}) \mathtt{T}$, composed of the incremental motion $\Delta\bm{\xi} \in \ensuremath{\mathfrak{se}(3)}\xspace$ and the current estimate $\mathtt{T} \in \ensuremath{\mathrm{SE}(3)}\xspace$. 
This allows to optimize the incremental update $\Delta\bm{\xi}$ in the tangent space of the current estimate $\mathtt{T}$.
Once the optimal increment $\Delta\bm{\xi}^*$ is found,  the transformation matrix $\mathtt{T}$ is updated as
\begin{equation}
    \label{eq:update_SE3}
    \mathtt{T} \leftarrow \mathrm{exp}_\mathrm{SE(3)}(\Delta\bm{\xi}^*) \mathtt{T}.
\end{equation}
Note that we follow the \textit{left-multiplication} convention to be consistent with \cite{strasdat2010scale,whelan2016elasticfusion}.

We refer the readers to \cite{corke2017robotics} for a reference on the representation of 6 DoF pose in the 3D space, and to \cite{strasdat2012local,sola2018micro} for introductions to Lie algebras for odometry and SLAM.

\section{Camera Tracking\index{camera tracking}}
\label{sec:motion}

In this section, we detail the algorithms that are most commonly used for estimating the 6 DoF motion of an RGB-D camera. The methods will be divided attending to the type of residual they minimize:


\begin{itemize}

    \item {\bf Methods based on photometric alignment (Section \ref{sec:directf2f}).} The alignment results from the minimization of a photometric error over corresponding pixels in two frames.

    \item {\bf Methods based on geometric alignment (Section \ref{sec:featuref2f}).} While direct methods minimize a photometric error, we refer to geometric alignment methods to those that minimize geometric residuals either in the image or 3D domains. 
    
\end{itemize}

Recent results suggest that direct methods present a higher accuracy than those based on geometric alignment, both in odometry \cite{engel2018direct} and mapping \cite{zubizarreta2019direct}. Most of the state-of-the-art systems are, because of this reason, based on dense frame alignment. Among the weaknesses of direct methods we can name their small basin of convergence, which can limit the accuracy in wide baselines cases, and their sensitivity to calibration errors, rolling shutter or unsynchronisation between the color and depth images \cite{schops2019bad}. 


\begin{figure}[t]
\centering
\includegraphics[width=\textwidth]{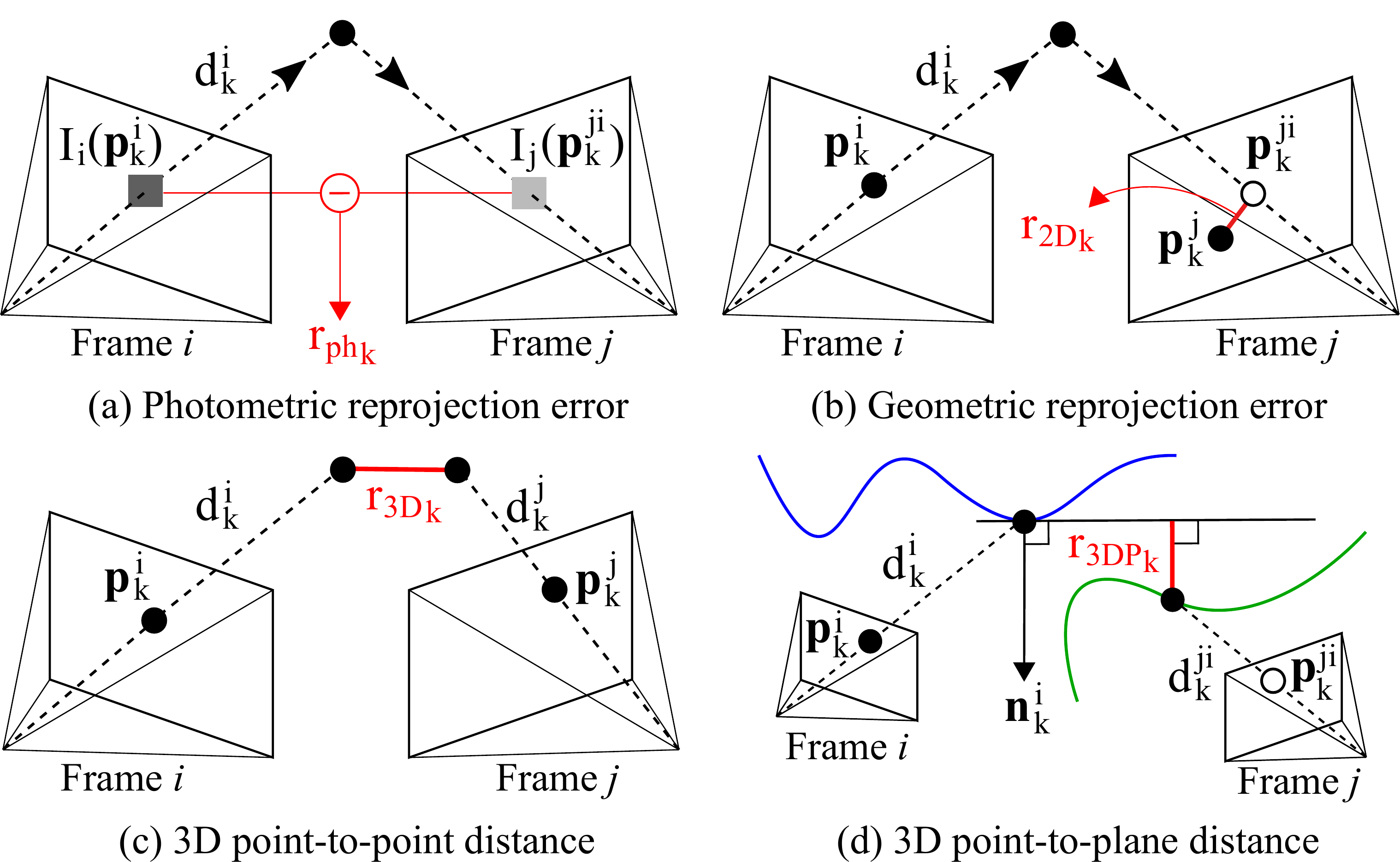}
\caption{
Different types of error criteria frequently used in the literature:
(a) A photometric reprojection error is the pixel intensity difference between a reference pixel in frame $i$ and its reprojection in frame $j$.
(b) Given a reference point in frame $i$, a geometric reprojection error is the image distance between its match and the reprojection in frame $j$.
(c) A 3D point-to-point distance is the Euclidean distance between the backprojections of two matched points.
(d) A 3D point-to-plane distance is the Euclidean distance between the tangent plane at the backprojected reference point in frame $i$ and the backprojected reprojection of the point in frame $j$.}
\label{fig:errors}
\end{figure}

\subsection{Photometric Alignment\index{photometric aligment}}
\label{sec:directf2f}


Assuming that the same scene point will have the same color in different images, photometric alignment aims to estimate the motion between two frames by minimizing the pixel intensity difference. 
This error criterion is called a \textit{photometric reprojection} error.
For each pixel ${\mathbf p}^i_k$ in the reference frame $i$, it is given by





\begin{equation}
    \label{ec:photores}
    {r_{ph}}_k({\Delta\bm{\xi}_{ji}}) =
    I_i\left({\mathbf p}^i_k\right)-
    I_j\big(\mathbf{p}^{ji}_{k}({\Delta\bm{\xi}_{ji}}) \ \big),
\end{equation}
where $\mathbf{p}^{ji}_{k}({\Delta\bm{\xi}_{ji}})$ is the reprojection of $\mathbf{p}^i_k$ in frame $j$ evaluated at the incrementally updated transformation $\mathrm{exp}_\mathrm{SE(3)}(\Delta\bm{\xi}_{ji})\mathtt{T}_{ji}$.
Fig. \ref{fig:errors}a illustrates this error.
Notice that in \eqref{ec:photores} we omitted some of the variables in the reprojection function $\mathbf{p}^{ji}_{k}$ for readability.
The full function is written as
\begin{equation}
    \label{eq:full_reprojection}
    \mathbf{p}^{ji}_{k}({\Delta\bm{\xi}_{ji}}) \stackrel{\eqref{eq:reprojection}}{=}\mathbf{p}^{ji}_{k}\left(\mathbf{p}^i_k, \ d^i_k, \  \mathrm{exp}_\mathrm{SE(3)}(\Delta\bm{\xi}_{ji})\mathtt{T}_{ji}\right).
\end{equation}

%

The total cost function to minimize, $E_{ph}$, is the weighted squared sum of the individual photometric errors for all considered pixels:

\begin{equation}
\label{eq:total_photometric}
\Delta\bm{\xi}_{ji}^* = \argmin_{\Delta\bm{\xi}_{ji}} \  E_{ph}(\Delta\bm{\xi}_{ji})
= \argmin_{\Delta\bm{\xi}_{ji}} \sum_k \omega({r_{ph}}_k)\left({r_{ph}}_k(\Delta\bm{\xi}_{ji})\right)^2
\end{equation}
with some weight function $\omega$, \emph{e.g.,} constant for unweighted least squares, or robust weight function such as Huber's \cite{huber2011robust}.
As discussed in Section \ref{sec:optimization}, this problem can be solved using IRLS. 
Once the optimal increment is found, $\mathtt{T}_{ji}$ is updated using \eqref{eq:update_SE3}, and this optimization process is iterated until convergence.

\cite{kerl2013robust} proposes a similar photometric alignment between consecutive frames of a video, achieving very accurate odometry results. 
For study of different alignment strategies, we refer to \cite{klose2013efficient}.
The photometric alignment can also be done in a frame-to-map basis. For example, in \cite{concha2017rgbdtam}, photometric and geometric errors are used to track the camera pose with respect to the closest keyframe in the map.

\subsection{Geometric Alignment\index{geometric aligment}}
\label{sec:featuref2f}



In contrast to photometric alignment that directly uses raw pixel intensities, geometric alignment estimates the camera motion by minimizing the Euclidean distances between the two corresponding sets of geometric primitives in 2D or 3D.

\textbf{2D Point-to-Point Alignment:} 
A \textit{geometric reprojection} error is the most representative type of 2D error used in VO and VSLAM.
This error is illustrated in Fig. \ref{fig:errors}b.
Given a point $\mathbf{p}^i_k$ in the reference frame $i$, it measures the image distance between its match $\mathbf{p}^j_k$ and the projection $\mathbf{p}^{ji}_k$ \eqref{eq:full_reprojection} in the current frame $j$:
\begin{equation}
    \label{eq:featurebasedframe2mapmotion}
    {r_{2D_{\scriptstyle k}}}(\Delta\bm{\xi}_{ji}) = \frac{\lVert\mathbf{p}^j_k-\mathbf{p}^{ji}_k(\Delta\bm{\xi}_{ji})\rVert}{\sigma^i_k} \quad \text{with} \quad \sigma^i_k = \left(\lambda_\text{pyr}\right)^{\displaystyle L_{\text{pyr},\mathbf{p}^i_k}},
\end{equation}
where $\sigma^i_k$ is the standard deviation of the image point $\mathbf{p}^i_k$ that depends on the scale factor of the image pyramid $\lambda_\text{pyr} (>1)$
and the level $L_{\text{pyr},\mathbf{p}^i_k}$ at which the point was detected.

Unlike photometric errors, geometric errors require data association. 
For sparse points, this can be done by matching feature descriptors (\emph{e.g.}, SIFT \cite{lowe2004distinctive}, SURF \cite{bay2006surf}, ORB \cite{rublee2011orb}) or extracting salient corners (\emph{e.g.}, Harris corner \cite{harris1988combined}, FAST \cite{rosten2006machine} or Shi-Tomasi \cite{shi1994good} features) and tracking them \cite{lucas1981iterative}.
Aggregating ${{r}_{2D_{\scriptstyle k}}}$ for every point $k$, we obtain the total cost function analogous to \eqref{eq:total_photometric}:
\begin{equation}
\label{eq:total_2Dpoint2point}
\Delta\bm{\xi}_{ji}^* = \argmin_{\Delta\bm{\xi}_{ji}} \  E_{2D}(\Delta\bm{\xi}_{ji})
= \argmin_{\Delta\bm{\xi}_{ji}} \sum_k \omega(r_{2D_{\scriptstyle k}})\left(r_{2D_{\scriptstyle k}}(\Delta\bm{\xi}_{ji})\right)^2.
\end{equation}
Minimizing this cost function to estimate the camera motion is called \textit{motion-only} bundle adjustment, and this method is used among others in ORB-SLAM2 \cite{mur2017orb} for tracking.



\textbf{3D Point-to-Point Alignment:} 
Instead of minimizing the reprojection error in 2D image space, one can also minimize the distance between the backprojected points in 3D space (see Fig. \ref{fig:errors}c).
The 3D errors can be defined over dense point clouds or sparse ones.
For the latter case, the first step should be the extraction and matching of the sparse salient points in the RGB channels. Henry et al. \cite{henry2010rgb}, for example, uses SIFT features \cite{lowe2004distinctive}, although others could be used.


Given two sets of correspondences in image $i$ and $j$, the 3D geometric error is obtained as
\begin{gather}
    {r_{3D_{\scriptstyle k}}}(\Delta\bm{\xi}_{ji}) =  \big\lVert{\mathbf{P}^j_k}'-\mathbf{P}^{ji}_k(\Delta\bm{\xi}_{ji})\big\rVert \label{eq:icp_feature_based} \\ 
    \text{with} \ \  
    \mathbf{P}^{ji}_k(\Delta\bm{\xi}_{ji}):=\pi_\text{3D}\left(\mathrm{exp}_\mathrm{SE(3)}(\Delta\bm{\xi}_{ji})\mathtt{T}_{ji}\pi^{-1}_\text{3D}\left({\mathbf{P}^{i}_k}'\right)\right) \label{eq:corresponding_backprojection},
\end{gather}
where ${\mathbf{P}^i_k}'$ and ${\mathbf{P}^j_k}'$ are the 3D points backprojected from the 2D correspondence $\mathbf{p}^i_k$ and $\mathbf{p}^j_k$ using \eqref{eq:backprojection}.
Aggregating ${{r}_{2D_{\scriptstyle k}}}$ for every point $k$, we obtain the total cost function analogous to \eqref{eq:total_photometric} and \eqref{eq:total_2Dpoint2point}:
\begin{equation}
\label{eq:total_3Dpoint2point}
\Delta\bm{\xi}_{ji}^* = \argmin_{\Delta\bm{\xi}_{ji}} \  E_{3D}(\Delta\bm{\xi}_{ji})
= \argmin_{\Delta\bm{\xi}_{ji}} \sum_k \omega(r_{3D_{\scriptstyle k}})\left(r_{3D_{\scriptstyle k}}(\Delta\bm{\xi}_{ji})\right)^2.
\end{equation}

For the case of dense cloud alignment, the standard algorithm is \emph{Iterative Closest Point (ICP)} \cite{besl1992method}.
ICP alternates the minimization of a geometric distance between points (the point-to-point distance in equation \ref{eq:icp_feature_based} or the point-to-plane one defined later in this section) and the search for correspondences (usually the nearest neighbours in the 3D space).

The strengths and limitations of sparse and dense cloud alignment are complementary for RGB-D data. Aligning dense point clouds can lead to more accurate motion estimation than aligning sparse ones, as they use more data. On the other hand, ICP might diverge if the initial estimate is not sufficiently close to the real motion. In practice, combining the two is a preferred approach: Sparse alignment, based on feature correspondences, can produce a robust and reliable initial seed. Afterwards, dense alignment can refine such initial estimate using ICP.

\textbf{3D Point-to-Plane Alignment:}  The point-to-plane distance, that minimizes the distance along the target point normal, is commonly used in dense RGB-D point cloud alignment \cite{henry2010rgb,newcombe2011kinectfusion,whelan2015elasticfusion_wo_posegraph,dai2017bundle}. 
The residual is in this case

\begin{equation}
\label{eq:icp_point2plane}
    r_{3DP_{\scriptstyle k}}
    (\Delta\bm{\xi}_{ji}) =
    \Biggr|
    \mathbf{n}^i_k\cdot
    \left(
    {\mathbf{P}^i_k}'-
    \left(
    \text{exp}_\text{SE(3)}(\Delta\bm{\xi}_{ji})\mathtt{T}_{ji}\right)^{-1}\left(d^{ji}_k\frac{\mathtt{K}^{-1}\pi^{-1}_\text{2D}\left(\mathbf{p}^{ji}_k(\Delta\bm{\xi}_{ji})\right)}{\lVert\mathtt{K}^{-1}\pi^{-1}_\text{2D}\left(\mathbf{p}^{ji}_k(\Delta\bm{\xi}_{ji})\right)\rVert}\right)
    \right)
    \Biggr|,
\end{equation}
where ${\mathbf{P}^i_k}'$ is the 3D backprojection of $\mathbf{p}^i_k$ using \eqref{eq:backprojection}, $\mathbf{n}_k^i$ is the surface normal at ${\mathbf{P}^i_k}'$, $\mathbf{p}^{ji}_k(\Delta\bm{\xi}_{ji})$ is the reprojection of $\mathbf{p}^i_k$ in frame $j$ evaluated at the incrementally updated transformation $\text{exp}_\text{SE(3)}(\Delta\bm{\xi}_{ji})\mathtt{T}_{ji}$, which is given by \eqref{eq:full_reprojection} and \eqref{eq:reprojection}, and $d^{ji}_k$ is the measured depth at this reprojection in frame $j$.
This error is illustrated in Fig. \ref{fig:errors}d.
Aggregating ${{r}_{3DP_{\scriptstyle k}}}$ for every point $k$, we obtain the total cost function analogous to \eqref{eq:total_photometric}, \eqref{eq:total_2Dpoint2point} and \eqref{eq:total_3Dpoint2point}:
\begin{equation}
\label{eq:total_3Dpoint2plane}
\Delta\bm{\xi}_{ji}^* = \argmin_{\Delta\bm{\xi}_{ji}} \  E_{3DP}(\Delta\bm{\xi}_{ji})
= \argmin_{\Delta\bm{\xi}_{ji}} \sum_k \omega(r_{3DP_{\scriptstyle k}})\left(r_{3DP_{\scriptstyle k}}(\Delta\bm{\xi}_{ji})\right)^2.
\end{equation}

\section{Scene Mapping\index{mapping}}
\label{sec:mapping}

In this section we briefly survey the main algorithms for estimating scene maps from several RGB-D views. There are two basic types of scene representations that are commonly used, that we will denote as \textbf{point-based maps (Section \ref{sec:pointcloudmap})} and \textbf{volumetric maps (Section \ref{sec:volmap})}. 

\subsection{Point-Based Mapping}
\label{sec:pointcloudmap}

Representing a scene as a set of points or surfels is one of the most common alternatives for estimating local maps of a scene. Bundle Adjustment \cite{triggs1999bundle}, consisting on the joint optimization of a set of camera poses and points, is frequently used to obtain a globally consistent model of the scene \cite{mur2017orb}. 
However, there are also several recent VSLAM approaches that alternate the optimization between points and poses, reducing the computational cost with a small impact in the accuracy, given a sufficient number of points \cite{zhou2014color,platinsky2017monocular,yokozuka2019vitamin,schops2019bad}.

In its most basic form, the map model consists of a set of $n$ points and $m$ RGB-D keyframes. 
Every point is represented by its 3D position in the world reference frame ${\mathbf P}_k^w$. For every keyframe $i$, we store its pose $\mathtt{T}_{iw}$ and its RGB-D image $I_i$.


Similarly to camera tracking in Section \ref{sec:motion}, map optimization algorithms are based on the photometric or geometric alignment between the keyframes. In this case, however, both the keyframe poses and point positions are optimized. 

\textbf{Photometric Bundle Adjustment\index{photometric bundle adjustment}:}
This method minimizes a cost function similar to \eqref{eq:total_photometric}, with the difference that it does not backproject the 2D points using the measured depths.
Instead, it aims to find the 3D point that minimizes the photometric errors in all keyframes where it was visible. 
Let $\mathbf{P}_{\mathcal{M}} = \left( \mathbf{P}_1, \hdots, \mathbf{P}_k, \hdots, \mathbf{P}_n \right)^\top$ be the set of all map points and $\Delta\bm{\xi}_{\mathcal{M}} = \left( \Delta\bm{\xi}_{1w}, \hdots, \Delta\bm{\xi}_{jw}, \hdots, \Delta\bm{\xi}_{mw} \right)^\top$ the set of incremental transformations to the current estimates of the keyframe poses.
Then, the optimization problem is formulated as
\begin{align}
\{\Delta\bm{\xi}_{\mathcal{M}}^*, \mathbf{P}^*_{\mathcal{M}}\}
&= \argmin_{\Delta\bm{\xi}_{\mathcal{M}},\mathbf{P}_{\mathcal{M}}} \  E_{ph}(\Delta\bm{\xi}_{\mathcal{M}},\mathbf{P}_{\mathcal{M}})\\
&= \argmin_{\Delta\bm{\xi}_{jw},\mathbf{P}^w_k} \sum_{j} \sum_{k} \omega({r_{ph}}_k)\left({r_{ph}}_k(\Delta\bm{\xi}_{jw},\mathbf{P}^w_k)\right)^2
\end{align}\vspace{-1em}
with
\begin{align*}
&{r_{ph}}_k(\Delta\bm{\xi}_{jw},\mathbf{P}^w_k) \nonumber\\
&=
\begin{cases}
0 & \text{if} \ \mathbf{P}_k \text{ is not visible} \\
 & \text{in frame } j,\\
I_i\left({\mathbf p}^i_k\right)-
I_j\big(\pi_\text{2D}\big(\mathtt{K}\pi_\text{3D}(\mathrm{exp}_\text{SE(3)}(\Delta\bm{\xi}_{jw})\mathtt{T}_{jw}\pi_\text{3D}^{-1}(\mathbf{P}^w_k))\big) \ \big) & \text{otherwise},
\end{cases}
\end{align*}
where $I_i\left({\mathbf p}^i_k\right)$ is the pixel intensity at which $\mathbf{P}_k^w$ was detected in its reference keyframe $i$ (\emph{i.e.,} the keyframe in which the point was first detected and parameterized).

\textbf{Geometric Bundle Adjustment\index{geometric bundle adjustment}:}
This method minimizes a cost function similar to \eqref{eq:total_2Dpoint2point}, with the difference that the reprojection with the measured depth is replaced by the projection of the current estimate of the 3D point. 
Using the same notation as for the photometric bundle adjustment, the optimization problem is formulated as
\begin{align}
\{\Delta\bm{\xi}_{\mathcal{M}}^*, \mathbf{P}^*_{\mathcal{M}}\}
&= \argmin_{\Delta\bm{\xi}_{\mathcal{M}},\mathbf{P}_{\mathcal{M}}} \  E_{2D}(\Delta\bm{\xi}_{\mathcal{M}},\mathbf{P}_{\mathcal{M}})\\
&= \argmin_{\Delta\bm{\xi}_{jw},\mathbf{P}^w_k} \sum_{j} \sum_{k} \omega({r_{2D}}_k)\left({r_{2D}}_k(\Delta\bm{\xi}_{jw},\mathbf{P}^w_k)\right)^2
\end{align}\vspace{-1em}
with
\begin{align*}
&{r_{2D}}_k(\Delta\bm{\xi}_{jw},\mathbf{P}^w_k) \nonumber\\
&=
\begin{cases}
0 &\text{if} \ \mathbf{P}_k \text{ is not detected} \\
 & \text{in frame } j,\\[-0.5em]
\displaystyle\frac{\bigg\lVert\mathbf{p}^j_k
-
\pi_\text{2D}\left(\mathtt{K}\pi_\text{3D}\left(\mathrm{exp}_\text{SE(3)}(\Delta\bm{\xi}_{jw})\mathtt{T}_{jw}\pi_\text{3D}^{-1}(\mathbf{P}^w_k)\right)\right)
\bigg\rVert}{\sigma^j_k} 
& \text{otherwise}.
\end{cases}
\end{align*}
Note that $\sigma^j_k$ is defined in \eqref{eq:featurebasedframe2mapmotion}.

\subsection{Volumetric Mapping}
\label{sec:volmap}

One of the main weaknesses of point-based representations for mapping is that they do not model the empty and occupied space. 
This can be a problem for applications such as robot navigation or occlusion modelling in AR. Volumetric mapping aims to overcome such problems by modeling the occupancy of the whole 3D scene volume. 

The most usual model for volumetric maps is the Truncated Signed Distance Function \cite{curless1996volumetric}, used for example in \cite{newcombe2011kinectfusion,whelan2013robust,whelan2015real,klingensmith2015chisel}.
In this representation, the 3D world is discretized into voxels and modeled as a volumetric signed distance field $\Phi:\mathbb{R}^3\rightarrow \mathbb{R}$, where we assign to each cell the distance to the nearest object, which is defined positive if its center is outside the object and negative if it is inside it.
Since only the surfaces and their surroundings are considered, the distances are usually truncated if larger than a threshold $\tau$. Also, for every cell, a weight is stored that represents the confidence on the distance measurement. The algorithm for updating a TSDF with new depth measurements measurement was first presented in \cite{curless1996volumetric}. In a few words, it consists on a weighted running average on the distance measurements from the depth sensors.

TSDF is addressed in depth in chapter~5 of this book.
For this reason, we do not extend further on it and refer the reader to this chapter, and the references there and in this section, for further detail on this topic.

\section{Loop Closing\index{loop closing}}
\label{sec:loopclosure}

Loop closing algorithms correct the drift that has accumulated during exploratory trajectories, maintaining a consistent global representation of the environment. \textbf{Loop detection (Section \ref{sec:placerecognition})}, is mainly based on the visual appearance between two keyframes of the map. When these two keyframes are imaging the same place and the loop closure has been detected, the geometric constraint between the two is added to the map, which is then updated according to it. This map update is known as \textbf{loop correction (Section \ref{sec:posegraphs})}, and we detail the pose graph formulation as an efficient alternative for large map representations and loop closing correction.

\subsection{Loop Detection\index{loop detection}}
\label{sec:placerecognition}

Due to the excellent performance of visual place recognition, many RGB-D SLAM systems use only the RGB channels for loop detection (\emph{e.g.}, \cite{mur2017orb,concha2017rgbdtam,gutierrez2018rgbid}). The most used approaches are based on the bag of words model, first proposed in \cite{sivic2003video}. The implementation in \cite{galvez2012bags} is particularly suited for visual SLAM, adding robustness to plain visual appearance by geometric and sequential consistency checks. 

In the bag of words model the space of local descriptors is divided into discrete clusters using the k-means algorithm. Each cluster is referred to as a visual word, and the set of all visual words forms a visual dictionary. With such a partition, an image is described as the histogram of visual word occurrences. The place querying can be made very efficient by maintaining inverse indexes from the visual words to the database images in which they appear. 

Bag-of-words descriptors have some limitations for RGB-D odometry and SLAM. They assume images of sufficient texture to extract salient point features, and they do not use the information of the depth channel from RGB-D images. Also, the extraction and description of local features has a considerable computational overhead. 

There are several approaches in the literature that overcome such limitations. \cite{gee20126d} proposes to find loop closure candidates without features, by the alignment of keyframes against synthetic views of the map. \cite{shotton2013scene} uses regression forests to predict correspondences between an RGB-D frame and the map, an approach that has been refined in \cite{guzman2014multi,valentin2015exploiting,cavallari2017fly} among others. \cite{glocker2015real} proposed to encode each RGB-D image using randomized ferns. 


\subsection{Loop Correction\index{loop correction}}
\label{sec:posegraphs}

\begin{figure}[t]
\centering
\includegraphics[width=\textwidth]{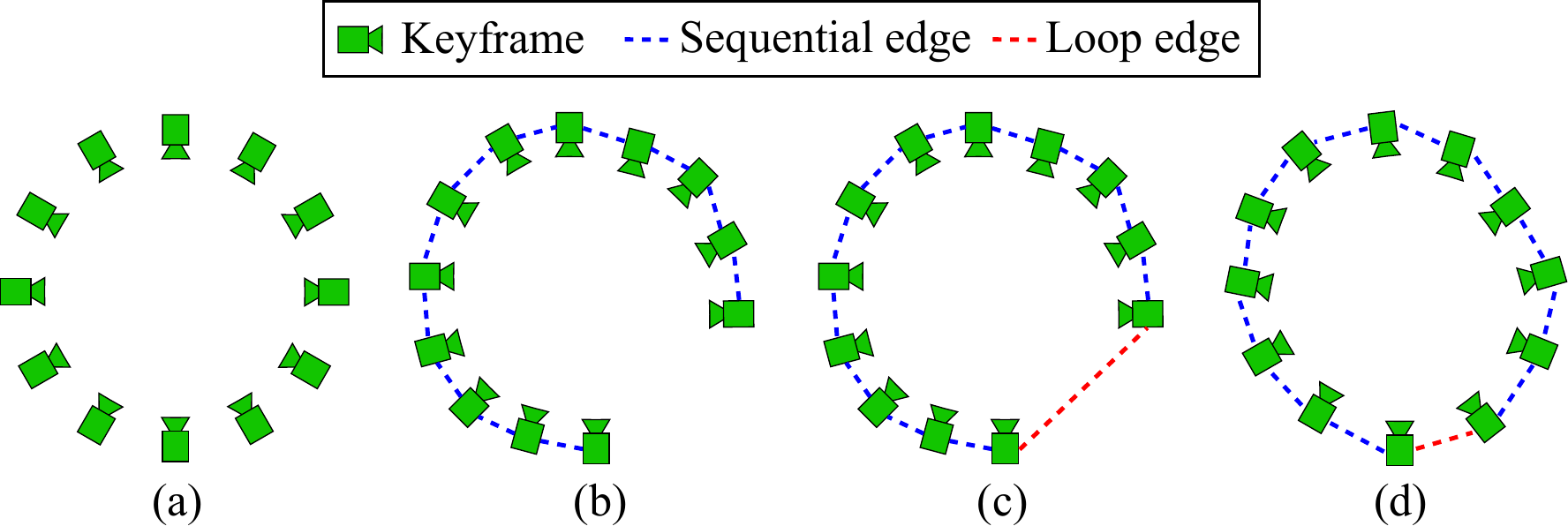}
\caption{An illustration of loop closure: (a) Ground truth. (b) Odometry result containing drift. (c) A loop detection followed by the computation of the loop constraint. (d) The keyframe trajectory after the pose graph optimization.}
\label{fig:loop_closing}
\end{figure}

Once a loop is detected based on the appearance of two keyframes, a constraint between the poses of both can be computed by photometric and/or geometric alignment. 
When such constraint is added to the map optimization, the global map consistency is achieved by accommodating this new constraint and correcting the accumulated drift.
For computational reasons, this correction is frequently done by pose graph optimization.
Fig. \ref{fig:loop_closing} illustrates a loop closure process.

A pose graph\index{pose graphs} is a compact map representation composed of the set of $m$ keyframe poses summarizing the trajectory, \emph{i.e.}, $\mathtt{T}_\text{kfs}=\{\mathtt{T}_{1a}, \mathtt{T}_{2a}, \hdots, \mathtt{T}_{ma}\}$ where the reference frame $a$ is chosen from one of the keyframes as the ``anchor'' to the rest.
As this representation does not include map points, it is particularly useful for estimating globally consistent maps of large areas at a reasonable cost, and is used among others in \cite{kerl2013dense,endres2012evaluation,concha2017rgbdtam}.


Pose graph optimization aims to minimize the following cost:
\begin{equation}
    \argmin_{\displaystyle\mathtt{T}_\text{kfs}} \ E_\text{graph} = \argmin_{\displaystyle\mathtt{T}_\text{kfs}} \sum_{(i,j)\in \epsilon_\text{edge}} \mathbf{r}_{ij}^\top \mathtt{\Omega}_{ij}^{-1} \mathbf{r}_{ij}
\end{equation}
where $\epsilon_\text{edge}$ denotes the set of edges (\emph{i.e.}, relative pose constraints) in the pose graph, $\mathbf{r}_{ij}$ and $\mathtt{\Omega}_{ij}$ are respectively the residual associated to the $i^{th}$ and $j^{th}$ camera poses and its uncertainty. Such residual is defined as 
\begin{equation}
    \mathbf{r}_{ij} = \mathrm{log}_\mathrm{SE(3)}(\mathtt{T}_{ij,0}\mathtt{T}_{ja}\mathtt{T}_{ia}^{-1})
    \label{eq:posegraphresidual}
\end{equation}
where $\mathtt{T}_{ij,0}$ is the fixed transformation constraint from the alignment (Section \ref{sec:motion}) and $\mathtt{T}_{ja}\mathtt{T}_{ia}^{-1}=\mathtt{T}_{ji}$ is the current estimate of the relative motion.
For more details on the pose graph optimization method, the reader is referred to \cite{kummerle2011g2o,rosen2019sesync}.

\section{Advanced topics}
\label{sec:references}

In this section, we review some of the relevant approaches in RGB-D odometry and SLAM that, due to space reasons, were not covered in the main part of the chapter.

\subsection{Hybrid Cost Function}
In Section \ref{sec:motion} and \ref{sec:mapping}, we discussed different types of cost functions separately.
Many state-of-the-art methods, however, minimize a weighted sum of multiple cost functions.
This strategy allows for better utilization of RGB-D data, which can lead to performance gains \cite{kerl2013dense,meilland2013super,damen2012egocentric,meilland2013unifying}.
In \cite{henry2010rgb}, 3D point-to-point error was used for outlier rejection, and then the pose was refined by minimizing the combined 2D point-to-point cost and 3D point-to-plane cost.
In \cite{whelan2015elasticfusion_wo_posegraph,dai2017bundle}, the joint minimization of photometric and point-to-plane cost was used.
Another popular method is to jointly minimize the photometric and (inverse) depth cost (which is not discussed here) \cite{kerl2013dense,steinbrucker2013large,babu2016sigma,gutierrez2015inverse,concha2017rgbdtam}.












\subsection{Semantic Mapping\index{semantic mapping}}
\begin{figure}[ht!]
\centering
\includegraphics[width=0.7\textwidth]{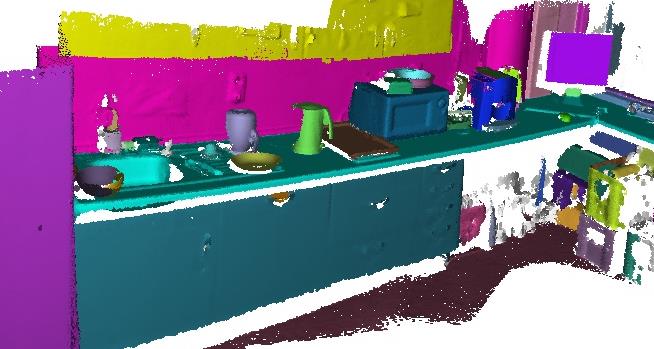}
\caption{Illustration of results from semantic RGB-D SLAM. Different colors indicate different object categories. Figures taken from \cite{tateno2015real}.}
\label{fig:semanticslam}
\end{figure}

In recent years, there has been an impressive progress in the field of machine learning (specifically deep learning) for visual recognition and segmentation tasks.
Building on them, there have appeared several visual SLAM algorithms that not only estimate geometric models, but also annotate them with high-level semantic information (see Fig. \ref{fig:semanticslam} for an illustration). 
The research on semantic mapping is not as mature as geometric mapping, with challenges related to robustness, accuracy and cost.
The state-of-the-art systems, however, show promising results.
Semantic mapping could improve the accuracy and robustness of current SLAM algorithms, and widen their applications. 
For example, \cite{bescos2018dynaslam} uses a combination of geometry and learning to remove dynamic objects and create life-long maps, achieving better accuracy than geometric SLAM baselines. Similarly, \cite{keller2013real} uses data association failures and region growing to segment and remove dynamic objects, improving the system robustness and accuracy.

One can differentiate between maps based on specific object instances and object categories. 
An approach like \cite{salas2013slam++} adopts the map of the former type.
It assumes that a database of relevant objects in the scene is available. 
The map is then modeled as a graph of keyframe and object poses, and it is optimized using the constraints from keyframe-to-keyframe point cloud alignment and keyframe-to-object using \cite{drost2010model}.
Object-based RGB-D SLAM has also been addressed in \cite{sunderhauf2017meaningful}

Most category-wise semantic mapping methods leverage 2D segmentation algorithms (\emph{e.g.}, \cite{he2017mask}), differing on how they transfer the labels to the 3D maps.
As a few examples, we refer to the following works for this area of research \cite{hermans2014dense,mccormac2017semanticfusion,ma2017multi,stuckler2015dense,pham2015hierarchical}.

\subsection{Edge-based methods\index{edge-based SLAM}} 
    While the majority of the existing methods consider each pixel as independent measurements, edge-based methods exploit the structural regularities of indoor scenes, modelling the scene geometry with lines or edges.
    This can provide an advantage over point-based methods, especially when the scene has weak texture but strong structural priors.
    One of the earliest works that demonstrated the advantage of edge-based registration in RGB-D SLAM is \cite{choi2013rgbd}.
    This method is based on an efficient edge detection for RGB-D point clouds and 3D registration of the edge points using the ICP algorithm.
    In \cite{bose2016fast}, it is shown that the edge detection can be accelerated using the previous RGB-D frame.
    On the other hand, \cite{lu2015robustness} proposes to model the straight lines only and incorporate their uncertainties in the pose estimation problem.
    Although this work is shown to outperform \cite{choi2013rgbd} under lighting variations, it fails when the scene contains few lines. 
    To overcome this limitation, \cite{lu2015robust} uses both points and lines.
    In \cite{kuse2016robust}, direct edge alignment is proposed that minimizes the sum of squared distances between the reprojected and the nearest edge point using the distance transform of the edge-map. 
    Other works propose to jointly minimize this edge distance and other errors, \emph{e.g.}, a photometric error \cite{wang2016edge} and an ICP-based point-to-plane distance \cite{schenk2017combining}.
    Later works such as \cite{zhou2019canny} and \cite{kim2018edge} take the image gradient direction also into account for the direct edge alignment.
    As in \cite{kerl2013robust}, these last two works estimate the camera pose using the iteratively reweighted least-squares (IRLS) method with the t-distribution as a robust weight function. 
    
\subsection{Plane-based methods\index{plane-based SLAM}} 
    Like edges, planes are abundant in man-made environments and can be modelled explicitly for tracking and mapping. 
    In \cite{taguchi2013point}, an RGB-D SLAM system is proposed based on the 3D registration between the minimal set of point/plane primitives.
    This system is improved in \cite{ataer2013tracking} and \cite{ataer2016pinpoint} by incorporating the guided search of points/planes and triangulation of 2D-to-2D/3D point matches, respectively.
    \cite{raposo2013plane} proposes an odometry method that uses planes (and points if strictly necessary) and refines the relative pose using a direct method.
    In \cite{salasmoreno2014dense}, a dense SLAM method is proposed based on dense ICP with a piecewise planar map.
    In both \cite{ma2016cpa} and \cite{hsiao17icra}, it is proposed to model planes in a global map, so that they are optimized together with the keyframe poses in the graph optimization for global consistency. 
    The main difference is that the former uses direct image alignment in an EM framework, while the latter combines geometric and photometric methods for the fast odometry estimation.
    Besides, the latter adopts the minimal plane parameterization proposed in \cite{kaess2015simultaneous} and does not require GPU.
    A visual-inertial method based on \cite{hsiao17icra} is proposed in \cite{hsiao2018dense}.
    In \cite{gao2015robust}, it is proposed to use planar point features for tracking and mapping, as they are more accurate than the traditional point features and computationally inexpensive.
    Other works such as \cite{le2017dense,kim2018low,kim2018linear} use Manhattan world assumption, which simplifies the incorporation of the planes into a SLAM formulation.
    Finally, \cite{proenca2018probabilistic} shows that it can be beneficial to use points, lines and planes all together in a joint optimization framework.

\subsection{Multisensor fusion\index{multisensor fusion}} 
The constraints coming from RGB-D data can be combined with other sources of information to increase the accuracy and robustness of the tracking and mapping processes. 
For example, \cite{laidlow2017dense} presents a tightly coupled formulation for RGB-D-inertial SLAM based on ElasticFusion \cite{whelan2016elasticfusion}. 
In \cite{klingensmith2016articulated}, RGB-D SLAM estimates the configuration space of an articulated arm. \cite{houseagoko} adds odometric and kinematic constraints from a wheeled robot with a manipulator, and \cite{scona2017direct} adds the kinematic constraints of a humanoid robot and inertial data. 

\subsection{Non-rigid reconstructions} 
The 3D reconstruction of non-rigid environments is a very relevant and challenging area of research that has been frequently addressed using RGB-D sensors. \cite{newcombe2015dynamicfusion} is one of the most representative systems, achieving impressive results for deformable surfaces. \cite{runz2018maskfusion} is a recent work that reconstructs a scene with multiple moving objects. 
\cite{jaimez2017fast} estimates very efficiently the odometry of an RGB-D camera and the flow of a scene that might contain static and dynamic parts.
\cite{scona2018staticfusion} classifies the scene parts into static and dynamic, fuses the static parts and discard the dynamic ones.
A recent survey on 3D reconstruction from RGB-D camera, including dynamic scenes, is conducted in \cite{zollhofer2018state}.
It places emphasis on high-quality offline reconstruction, which is complementary to the focus of this chapter on real-time online reconstruction and camera tracking.

\section{Conclusions}
\label{sec:conclusions}

Estimating the camera ego-motion and the 3D structure of the surrounding environment is a crucial component in many applications such as photogrammetry, AR and vision-based navigation.
For this particular tasks, RGB-D cameras provide significant advantages over RGB cameras, as the additional depth measurements ease the process of metric scene reconstruction.  
Furthermore, they impose a relatively mild constraint on cost, size and power, making them a popular choice for mobile platforms.
As a result, both academia and industry have shown an ever-increasing interest in RGB-D odometry and SLAM methods for the past decade.

In this chapter, we reviewed the general formulations of RGB-D odometry and SLAM.
The standard pipeline of VSLAM systems consists of three main components: camera pose tracking, scene mapping and loop closing.
For tracking and mapping, we discussed some of the widely-used methods and highlighted the difference in their formulations (\emph{i.e.}, photometric vs. geometric alignment and point-based vs. volumetric mapping).
For loop closing, we detailed the underlying principles of loop detection and drift correction, namely the appearance-based place recognition and pose graph optimization.
Lastly, we presented a brief review of the advanced topics in the research field today.

\section{Resources}
\label{sec:resources}

There are a high number of available resources in the web related to RGB-D odometry and SLAM. We will refer here the most relevant open-source software and public databases.

\subsection*{Code}

\noindent \textbf{FOVIS} \cite{huang2017visual} (\url{https://fovis.github.io/}) \newline Implementation of a feature-based RGB-D odometry.

\noindent \textbf{DVO\_SLAM} \cite{steinbrucker2011real,kerl2013robust,kerl2013dense,kerl2015dense} (\url{https://github.com/tum-vision/dvo_slam}) \newline Implementation of a frame-to-frame RGB-D visual Odometry.

\noindent \textbf{RGBDSLAM\_v2} \cite{endres20143} (\url{https://github.com/felixendres/rgbdslam_v2}, \url{http://wiki.ros.org/rgbdslam},  \url{https://openslam-org.github.io/rgbdslam.html}) \newline Implementation of an RGB-D SLAM system, with a feature-based camera tracking and a pose graph as map model.

\noindent \textbf{ElasticFusion} \cite{whelan2016elasticfusion} (\url{https://github.com/mp3guy/ElasticFusion}) \newline RGB-D scene-centered SLAM system that models the scene as a set of surfels that are deformed to accommodate loop closures.

\noindent \textbf{RGBDTAM} \cite{concha2017rgbdtam} (\url{https://github.com/alejocb/rgbdtam}) \newline RGB-D SLAM system with a pose graph as map model and frame-to-frame tracking.

\noindent \textbf{MaskFusion} \cite{runz2018maskfusion} (\url{https://github.com/martinruenz/maskfusion}) \newline A recent semantic (object-based) RGB-D SLAM system for dynamic scenes.

\noindent \textbf{PlaneMatch}\cite{shi2018planematch} (\url{https://github.com/yifeishi/PlaneMatch}) \newline RGB-D SLAM algorithm that proposes a novel descriptor for planar surfaces and exploits correspondences between them. 

\subsection*{Databases}

\noindent \textbf{RGB-D SLAM Dataset and Benchmark (Also known as \emph{the TUM dataset})}\cite{sturm12iros} (\url{https://vision.in.tum.de/data/datasets/rgbd-dataset}). \newline It contains indoor recordings with ground truth camera pose in a wide variety of conditions: rotation-only and general motion, static and dynamic environments and small and mid-size scene coverage. 

\noindent \textbf{The ETH3D dataset}\cite{schops2019bad} (\url{https://www.eth3d.net/}). \newline A benchmark dataset for RGB-D SLAM (among others), recorded with synchronized global shutter cameras.

\noindent \textbf{The Matterport dataset}\cite{Matterport3D} (\url{https://github.com/niessner/Matterport}). \newline Annotated data captured throughout 90 properties with a Matterport Pro Camera.

\noindent \textbf{Scannet}\cite{dai2017scannet} (\url{http://www.scan-net.org/}). \newline RGB-D video dataset annotated with 3D camera poses, reconstructions, and instance-level semantic segmentations.

\noindent \textbf{The ICL-NUIM dataset}\cite{handa2014benchmark} (\url{https://www.doc.ic.ac.uk/~ahanda/VaFRIC/iclnuim.html}). \newline This dataset contain RGB-D sequences on synthetic scenes; hence with camera pose and scene ground truth. 

\noindent \textbf{InteriorNet}\cite{InteriorNet18} (\url{https://interiornet.org/}). \newline Dataset containing RGB-D-inertial streams for synthetic large scale interior scenes.


\bibliographystyle{spmpsci}


\end{document}